\documentclass[lettersize,journal]{IEEEtran}
\usepackage{amsmath,amsfonts}
\usepackage{algorithmic}
\usepackage{algorithm}
\usepackage{array}
\usepackage[caption=false,font=normalsize,labelfont=sf,textfont=sf]{subfig}
\usepackage{textcomp}
\usepackage{stfloats}
\usepackage{url}
\usepackage{verbatim}
\usepackage{graphicx}
\usepackage{cite}
\usepackage{makecell}
\usepackage{pifont}
\usepackage{xcolor}
\usepackage{booktabs}
\usepackage{amssymb}

\usepackage{makecell}
\definecolor{mgreen}{RGB}{1,150,74}

\usepackage{pifont}

\definecolor{custom_green}{RGB}{0, 175, 0}

\usepackage{xcolor}
\usepackage{colortbl}
\usepackage{multirow}

\begin{document}

\title{DAOcc: 3D Object Detection Assisted Multi-Sensor Fusion for \\ 3D Occupancy Prediction}

\author{Zhen Yang*\thanks{*Corresponding author (yangzhen1324@163.com)} \quad
Yanpeng Dong \quad
Jiayu Wang \quad
Heng Wang \quad
Lichao Ma \quad
Zijian Cui \quad
Qi Liu \quad
Haoran Pei \quad
Kexin Zhang \quad
Chao Zhang
\\
Beijing Mechanical Equipment Institute, Beijing, China
}

\markboth{Journal of \LaTeX\ Class Files,~Vol.~14, No.~8, August~2021}%
{Shell \MakeLowercase{\textit{et al.}}: A Sample Article Using IEEEtran.cls for IEEE Journals}

\IEEEpubid{
        \begin{minipage}{\textwidth}
                Copyright © 2025 IEEE. Personal use of this material is permitted. \\
                However, permission to use this material for any other purposes must be obtained from the IEEE by sending an email to pubs-permissions@ieee.org.
        \end{minipage}
}

\maketitle

\begin{abstract}
Multi-sensor fusion significantly enhances the accuracy and robustness of 3D semantic occupancy prediction, which is crucial for autonomous driving and robotics. However, most existing approaches depend on high-resolution images and complex networks to achieve top performance, hindering their deployment in practical scenarios. Moreover, current multi-sensor fusion approaches mainly focus on improving feature fusion while largely neglecting effective supervision strategies for those features. 
To address these issues, we propose \textbf{DAOcc}, a novel multi-modal occupancy prediction framework that leverages 3D object \textbf{d}etection supervision to \textbf{a}ssist in achieving superior performance, while using a deployment-friendly image backbone and practical input resolution. 
In addition, we introduce a BEV View Range Extension strategy to mitigate performance degradation caused by lower image resolution. 
Extensive experiments demonstrate that DAOcc achieves new state-of-the-art results on both the Occ3D-nuScenes and Occ3D-Waymo benchmarks, and outperforms previous state-of-the-art methods by a significant margin using only a ResNet-50 backbone and 256×704 input resolution. With TensorRT optimization, DAOcc reaches 104.9 FPS while maintaining 54.2 mIoU on an NVIDIA RTX 4090 GPU. 
Code is available at \url{https://github.com/AlphaPlusTT/DAOcc}.
\end{abstract}
\section{Introduction}
\label{sec:intro}

3D semantic occupancy prediction (occ)  is a critical task in autonomous driving~\cite{yan2021sparse, cao2022monoscene,wang2024panoocc} and robotic systems~\cite{wang2021learning, huang2023voxposer, wang2024embodiedscan}, where accurately understanding the environment is essential for safe and efficient navigation. Reliable occupancy prediction requires not just accurate spatial data but also a comprehensive understanding of the environment's context. Achieving this necessitates the integration of data from multiple sensors. LiDAR provides precise 3D spatial information for obstacle detection, while cameras capture visual details like color and texture for a deeper understanding of the scene. By combining these complementary data sources, the accuracy and robustness of occupancy prediction are significantly enhanced.

\begin{figure}[t]
    \centering
    \includegraphics[width=\columnwidth]{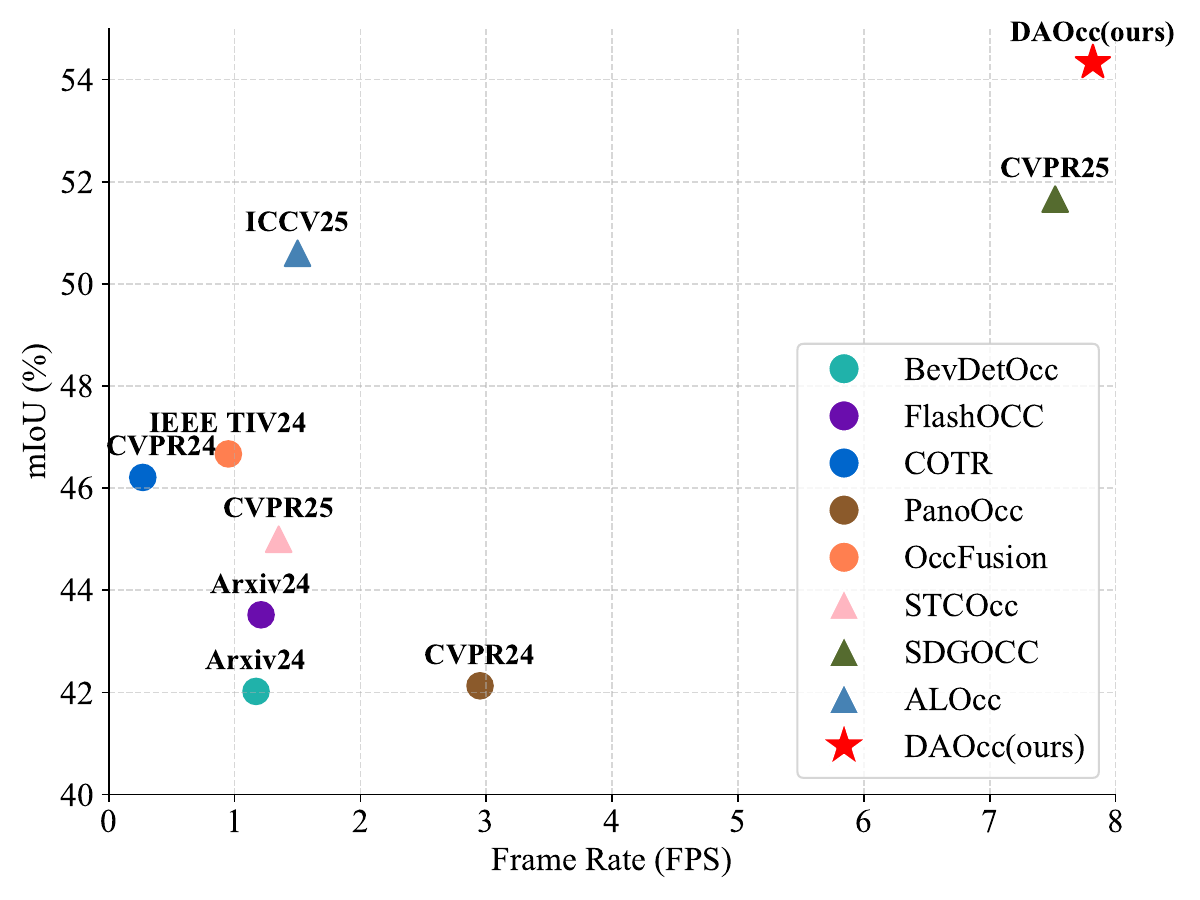}
    \caption{\textbf{Comparison of mIoU and FPS.} All methods are evaluated on the Occ3D-nuScenes validation set with the camera mask used during training. FPS for SDGOCC~\cite{duan2025sdgocc} and ALOcc~\cite{chen2024alocc} are sourced directly from their original publications, evaluated on NVIDIA RTX 4090. All other FPS results are obtained through our own benchmarking on identical hardware (NVIDIA RTX 4090) using the PyTorch FP32 backend. Our method, DAOcc, achieves the best trade-off, significantly outperforming previous works in both accuracy and real-time performance.}
    \vspace{-8pt}
    \label{fig:fps}
\end{figure}

In existing works on multi-modal~\cite{zhang2024radocc, ming2024occfusion, zhang2024occfusion, wang2023openoccupancy, pan2024co} or image-based~\cite{cao2022monoscene, wei2023surroundocc, tian2024occ3d, huang2021bevdet, wang2024panoocc, li2023fb, yu2023flashocc, shi2024occupancy, wu2024deep} occ tasks, achieving superior performance often involves using extremely high image resolutions and complex image feature extraction networks (see Table~\ref{tab:occ3d_nus_mask_w_fps}),
such as using 900×1600 resolution input image and ResNet101~\cite{he2016deep} equipped with DCN~\cite{dai2017deformable, zhu2019deformable}. However, this approach significantly limits the deployment of top-performing occ models on edge devices due to the high computational demands. In comparison to images, the point cloud is much sparser. For example, in the training set of the nuScenes~\cite{caesar2020nuscenes} dataset, the maximum number of points in a single point cloud frame is only 34,880, which is equivalent to just 2.4\% of the number of pixels in a 900×1600 resolution image. Therefore, how to effectively leverage point cloud data within a multi-modal occ framework remains to be further explored.

Moreover, we observe that most works~\cite{zhang2024radocc, ming2024occfusion, zhang2024occfusion, wang2023openoccupancy} on multi-modal occ primarily focuses on obtaining more effective fusion features, with insufficient exploration into the forms of supervision for these fused features. Although CO-Occ~\cite{pan2024co} introduces a regularization based on implicit volumetric rendering to supervise fused features, it only utilizes distance ground truth from the original point cloud data, failing to fully exploit the geometry and structure information inherent in point cloud features. In contrast, LiDAR-based 3D detectors~\cite{zhan2023real, chen2023focalformer3d, liu2024lion} effectively leverage this information, achieving significantly better performance in 3D object detection tasks compared to image-based 3D detectors~\cite{lin2023sparse4d, jiang2024far3d, liu2023sparsebev}. This observation suggests a new research direction: how to better exploit the unique strengths of point cloud data in multi-modal occ tasks.

\IEEEpubidadjcol
Based on these observations, we propose \textbf{DAOcc}, a novel multi-modal occ framework that leverages 3D object \textbf{d}etection to \textbf{a}ssist in achieving superior performance while using a deployment-friendly image encoder and practical input image resolution.

In constructing the baseline network of DAOcc, we adopt the simplest and most direct approach. First, we use a 2D image encoder and 3D sparse convolutions~\cite{yan2018second} to extract features from images and point clouds, respectively. Given that depth estimation from monocular images is an ill-posed problem~\cite{huang2023detecting, li2023bevstereo} and that deformable attention modules are overly complex~\cite{harley2023simple}, we employ a simple method similar to Harley \textit{et al.}~\cite{harley2019learning} to transform image features from 2D space to 3D volumetric space. Specifically, we project a set of predefined 3D points onto the 2D image feature plane and use bilinear interpolation to sample the corresponding 2D image features for these points. Next, we adopt a straightforward fusion strategy by concatenating the image and point cloud features and then applying 2D convolution to obtain a unified BEV (Bird’s Eye View) feature. Finally, we apply a fully convolutional BEV encoder with a residual structure to further fuse the unified BEV features, and subsequently employ the Channel-to-Height~\cite{yu2023flashocc} operation to transform the channel dimension into the height dimension.

\begin{figure}[t]
    \centering
    \includegraphics[width=\columnwidth]{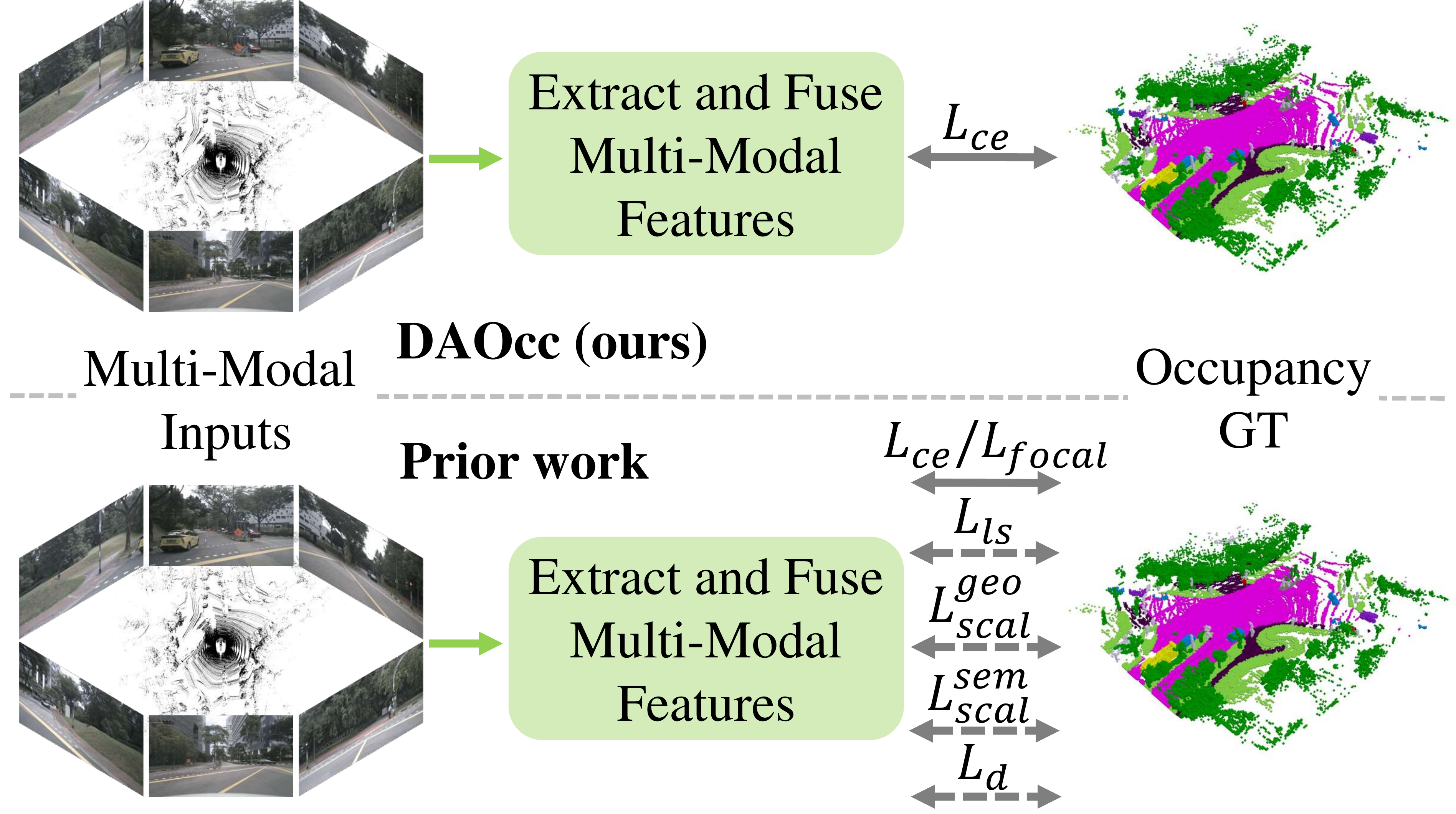}
    \caption{\textbf{Comparison of loss functions.} Unlike existing multi-modal methods~\cite{wang2023openoccupancy,ming2024occfusion, zhang2024occfusion, pan2024co}, which rely on complex combinations of loss functions, including several or all of the focal loss $\mathcal{L}_{focal}$~\cite{ross2017focal}, the scene-class affinity loss $\mathcal{L}_{scal}^{geo}$ and $\mathcal{L}_{scal}^{sem}$~\cite{cao2022monoscene}, the lovasz-softmax loss $\mathcal{L}_{ls}$~\cite{berman2018lovasz}, and the depth loss $\mathcal{L}_{d}$~\cite{li2023bevdepth}, to achieve optimal performance, our approach only requires a single cross-entropy loss $\mathcal{L}_{ce}$ for the occupancy prediction task.}
    \vspace{-8pt}
    \label{fig:loss}
\end{figure}

To fully exploit the geometry and structure information inherent in point cloud features, we enhance the baseline model's unified BEV features by incorporating 3D object detection supervision, thereby improving the discriminability of the unified BEV features.
The detection-to-occupancy supervision injects explicit object-localization cues into the occupancy pipeline, sharpening object boundaries and boosting foreground class performance while keeping inference complexity low---since the detection branch is used only during training and removed at inference. It also eliminates the need for complex loss function combinations, as depicted in Figure~\ref{fig:loss}.
Additionally, given the sparsity of point clouds, we extend the processing range of the point clouds and employ sparse convolutions~\cite{yan2018second} to mitigate the computational overhead introduced by this extension. 
To address the misalignment between the expanded point cloud range and the occupancy label, we apply a combination of coordinate system transformation and feature interpolation to extract fusion features that are spatially aligned with the occupancy label.
We refer to this approach as BVRE (BEV View Range Extension). BVRE provides a larger BEV field of view, offering more contextual information and mitigating the adverse effects of reduced image resolution. 

Extensive experiments are conducted to validate the efficiency and effectiveness of the proposed DAOcc. The results show that DAOcc establishes the new state-of-the-art performance on the Occ3D-nuScenes~\cite{tian2024occ3d} and Occ3D-Waymo~\cite{tian2024occ3d} benchmarks, while using only ResNet50~\cite{he2016deep} with a 256×704 input image resolution.
Specifically, on the Occ3D-nuScenes validation set, DAOcc achieves a 2.67 mIoU improvement over SGDOCC~\cite{duan2025sdgocc} when trained with camera mask and a 4.7 RayIoU improvement over ALOcc~\cite{chen2024alocc} without camera mask. Additionally, on the Occ3D-Waymo validation set, DAOcc demonstrates a 6.08 mIoU boost compared to BEVFormer-Fusion~\cite{tian2024occ3d}. Furthermore, with TensorRT optimization, DAOcc reaches 104.9 FPS on an NVIDIA RTX 4090 GPU while maintaining 54.2 mIoU on the Occ3D-nuScenes validation set, significantly surpassing the mIoU of previous state-of-the-art methods.

In conclusion, our contributions are summarized as follows:

\begin{itemize}
    \item We design a simple yet efficient multi-modal 3D semantic occupancy prediction baseline, eliminating the need for complex deformable attention~\cite{li2022bevformer} modules as well as image depth estimation during feature fusion.
    \item We propose DAOcc, a novel multi-modal occupancy prediction framework that leverages detachable 3D object detection supervision to improve foreground class discrimination and maintain low inference complexity, thus achieving superior performance while using a deployment-friendly image encoder and practical input image resolution.
    \item We introduce a BEV View Range Extension strategy to mitigate the adverse effects of reduced image resolution, and design a flexible feature processing pipeline consisting of coordinate system transformation and feature interpolation, which eliminates the complex manual feature space alignment.
    \item Our proposed method, DAOcc, establishes the new state-of-the-art performance on the Occ3D~\cite{tian2024occ3d} benchmark, while maintaining competitive inference speed. With TensorRT optimization, DAOcc can achieve 104.9 FPS while significantly surpassing the mIoU of the previous state-of-the-art methods.
\end{itemize}

\section{Related Work}
\label{sec:related work}

\subsection{3D Occupancy Prediction}
3D occupancy prediction aims to map all occupied voxels in the environment and assign semantic labels, thus providing more fine-grained perception results. 
While camera-based 3D occupancy prediction~\cite{cao2022monoscene,huang2023tri,zhang2023occformer,wang2023openoccupancy,wei2023surroundocc,tian2024occ3d,wang2024panoocc,yu2023flashocc,liu2023fully,ma2024cotr,liao2025stcocc,chen2024alocc, huang2021bevdet, shi2024occupancy, wu2024deep, li2023fb} has demonstrated promising results, multi-modal approaches offer greater reliability and robustness, making them indispensable for practical applications in autonomous driving and robotics. Due to the fact that cameras are susceptible to changing lighting and weather conditions, OccFusion~\cite{ming2024occfusion} enhances the accuracy and robustness of the occupancy network by integrating features from LiDAR and radar. In concurrent work, OccFusion~\cite{zhang2024occfusion} projects the preprocessed denser and more uniform point cloud onto the image plane to establish a mapping relationship, and performs deformable attention~\cite{zhu2020deformable} to fuse the corresponding features. Although OccFusion~\cite{zhang2024occfusion} avoids depth estimation, using deformable attention incurs a greater computational burden~\cite{harley2023simple}. 
Co-Occ~\cite{pan2024co} employs K-nearest neighbor search in selecting neighboring camera features to enhance the corresponding LiDAR features and proposes a regularization based on implicit volume rendering. However, this regularization only utilizes the distance ground truth of the point cloud and does not leverage its intrinsic geometry information. 
SDGOCC~\cite{duan2025sdgocc} leverages both the geometric and semantic information of point clouds to guide the 2D-3D view transformation, and introduces a fusion-to-occupancy-driven active distillation module that selectively transfers multimodal knowledge to image features based on LiDAR-identified regions. 
Compared to Co-Occ, SDGOCC additionally introduces point cloud semantic information, but its utilization of point cloud geometry information remains limited---relying solely on sparse point cloud based depth supervision for monocular depth estimation, without addressing the inherent ill-posed problem of this task.
In this work, we introduce a simple yet efficient multi-modal occupancy prediction network, eliminating the need for complex deformable attention~\cite{li2022bevformer} as well as image depth estimation during feature fusion, and avoiding the requirement for 3D object detection pre-training. 
Furthermore, to fully exploit the inherent geometry information in point cloud features, we incorporate 3D object detection as auxiliary supervision on the fused BEV features.

\subsection{Multi-Modal 3D Object Detection}
Multi-modal 3D object detection has achieved great success, surpassing camera-based~\cite{chen2024dsc3d,tao2023pseudo} and Lidar-based~\cite{li2023tinypillarnet} methods by leveraging semantic and geometric information. Recent multi-modal 3D object detection methods~\cite{liu2023bevfusion, li2022deepfusion, chen2023futr3d, wang2021pointaugmenting} mainly focus on learning effective BEV feature representations. TransFusion~\cite{bai2022transfusion} proposes a two-stage transformer-decoder based detection head and applies cross attention to obtain image features for each object query. BEVFusion~\cite{liu2023bevfusion} proposes an efficient and generic multi-task multi-sensor fusion framework which unifies multi-modal features in the shared BEV representation space, and introduces a specialized kernel to speed up the BEV pooling operation. In concurrent work, BEVFusion~\cite{liang2022bevfusion} decomposes the LiDAR-camera fusion into two streams that can independently output perception results, and performs feature fusion after the two streams. DAL~\cite{huang2023detecting} follows the concept of `Detecting As Labeling' and decouples the fused features during classification and regression. Specifically, it employs the fused features for classification, whereas it utilizes the point cloud features exclusively for regression. In this work, we incorporate the simple yet effective feature fusion approach of BEVFusion~\cite{liu2023bevfusion} and introduce 3D object detection as an auxiliary branch during training.

\subsection{Detection-Occupancy Interaction}
Recent work has explored tight couplings between 3D occupancy estimation and 3D object detection, with mutual benefits. OccupancyM3D~\cite{peng2024learning} uses voxel-wise occupancy as auxiliary dense supervision to enhance monocular 3D detection via richer 3D features and geometry priors. SOGDet~\cite{zhou2024sogdet} adds a semantic-occupancy branch that provides dense scene context to multi-view detectors, boosting localization and robustness. Zheng \textit{et al.}~\cite{zheng2024towards} recently proposed an object-centric occupancy completion approach, focusing on high-resolution modeling of foreground objects. 
DAOcc is complementary to these lines of work, while differing in both supervision direction and practical emphasis.

\section{Proposed Method}
\label{sec:proposed method}

\begin{figure*}[tb]
    \centering
    \includegraphics[width=\textwidth]{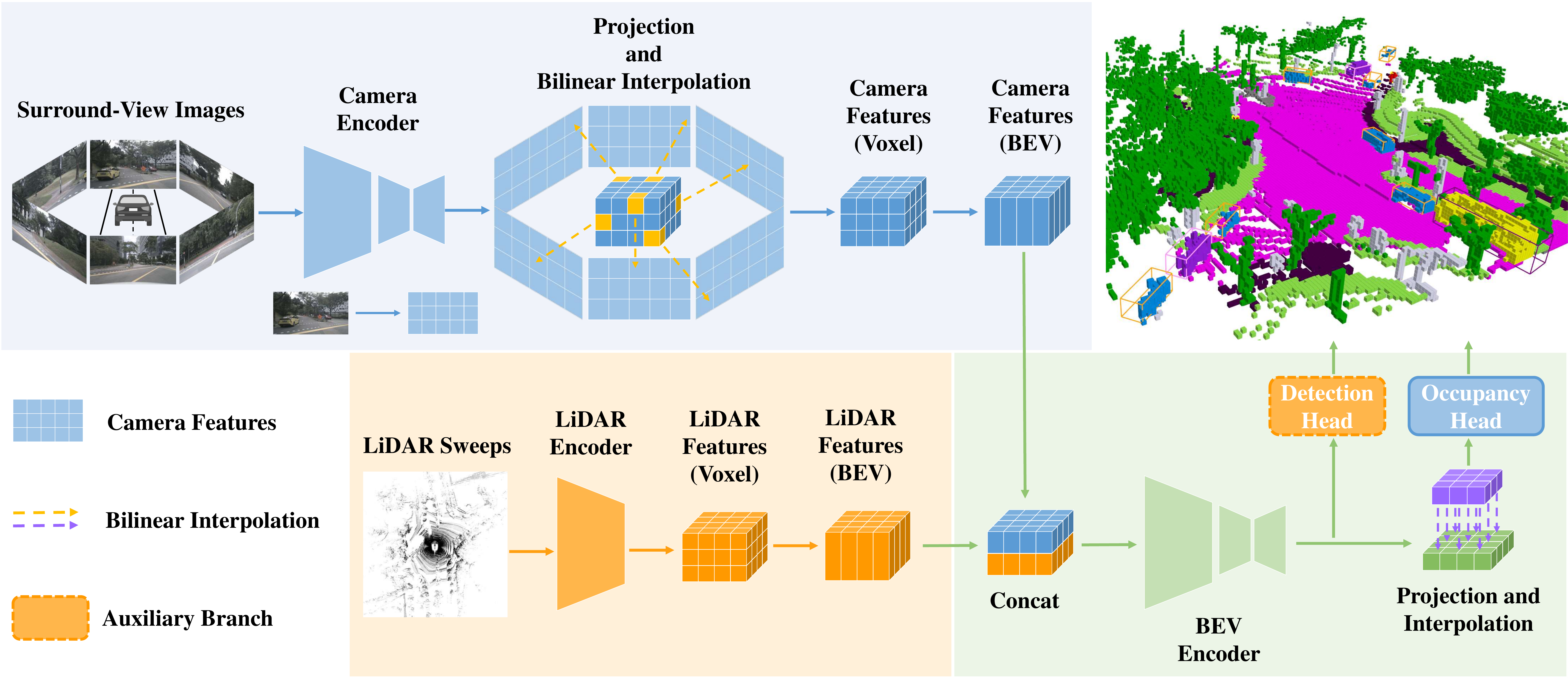}
    \caption{\textbf{Overview of our proposed DAOcc.} We first introduce the BVRE strategy to enrich the spatial contextual information by broadening the perceptual range from the BEV perspective. Then feature extraction is performed through the multi-modal occupancy prediction network illustrated in the above figure. Furthermore, to fully leverage the inherent geometry and structure information within point cloud features, we incorporate 3D object detection as auxiliary supervision.}
    \label{Fig:overview}
    \vspace{-8pt}
\end{figure*}

\begin{figure}[t]
    \centering
    \includegraphics[width=\columnwidth]{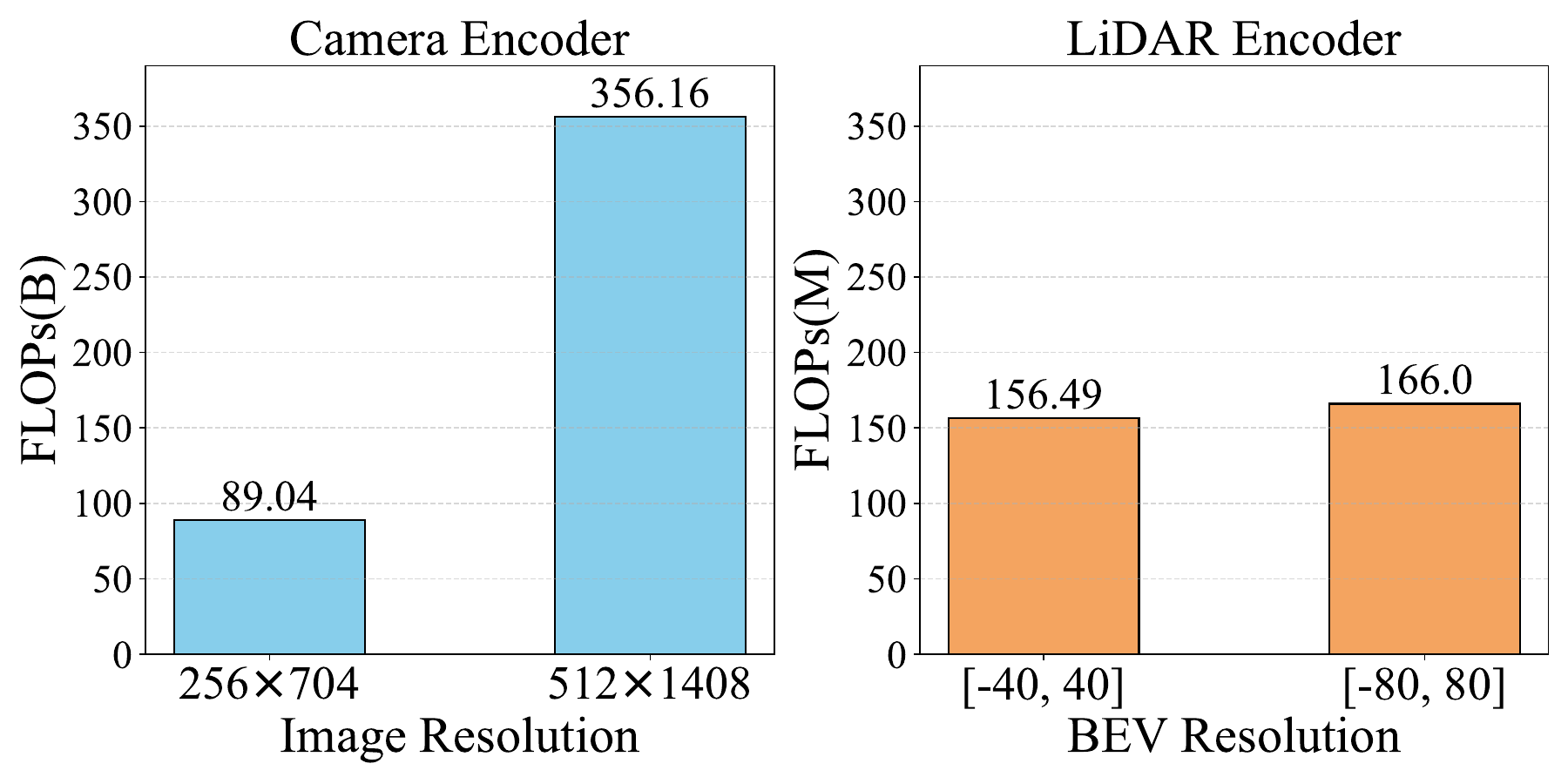}
    \caption{\textbf{Comparison of the impact of resolution increase on computational cost.} For LiDAR Encoder, the processing range of the point cloud along the $z$-axis in all experiments is from -5m to 3m. The example in the figure above selects the point cloud with the highest point count in the validation set and aggregates a total of 10 frames.}
    \vspace{-8pt}
    \label{fig:flops}
\end{figure}

\subsection{Overall Framework}

Our purpose is to fully utilize the inherent geometry and structure information in point cloud features within a multi-modal occupancy prediction framework. Previous multi-modal methods have not fully exploited this and can only achieve superior performance through a more complex image encoder and a larger input image resolution. The overall framework of our proposed DAOcc is illustrated in Figure~\ref{Fig:overview}. DAOcc takes surround images and their corresponding time-synchronized point cloud as input, and obtains the features of the image and point cloud through the Camera Encoder and LiDAR Encoder, respectively. The 2D image features are transformed into the 3D voxel space by projection and interpolation. Subsequently, the image and point cloud features in the 3D space are compressed along the height dimension to generate the corresponding BEV(Bird’s Eye View) features. A simple 2D convolution is then applied for feature fusion, and the fully convolutional BEV Encoder encodes these fused features to obtain the final BEV representations. Finally, the occupancy head uses the Channel-to-Height operation~\cite{yu2023flashocc} to restore the height of the BEV representations, resulting in final 3D voxel space representations that can be utilized for occupancy prediction. These modules collectively form the basic network architecture of DAOcc, which will be elaborated in Sec.~\ref{sec:basic network}.

On the foundation of the basic network, we introduce the BVRE strategy (Sec.~\ref{sec:bvre}) to compensate for the information loss that arises from the reduction in image resolution. This strategy aims to enrich the spatial contextual semantic information by broadening the perceptual range from the BEV perspective. Furthermore, to fully leverage the inherent geometry and structure information within point cloud features, we incorporate 3D object detection as auxiliary supervision (Sec.~\ref{sec:det head}). This auxiliary supervision not only enhances the discriminability of the fused features but also leads to a very concise overall training loss (Sec.~\ref{sec:total loss}) for our proposed framework.

\subsection{Basic Network}
\label{sec:basic network}

\subsubsection{LiDAR Encoder}
The method of embedding raw LiDAR points into 3D voxelized features is consistent with SECOND~\cite{yan2018second}. We first voxelize the point cloud, retaining a maximum of 10 points per voxel, resulting in a 3D voxel grid of size $D \times H \times W$.The feature representation for each voxel is obtained by averaging the features of all points within it. Next, we apply 3D sparse convolutions~\cite{yan2018second} to encode these voxel features, generating spatially compressed LiDAR voxel features $F_l \in \mathbb{R}^{C \times \frac{D}{16} \times \frac{H}{8} \times \frac{W}{8}}$, where $C$ represents the feature dimensions.

\subsubsection{Camera Encoder}
For image feature extraction, taking surround images as input, we first use ResNet50~\cite{he2016deep} as the backbone to extract multi-scale features, denoted as $F_{ms} = \{F_{\frac{1}{8}}, F_{\frac{1}{16}}, F_{\frac{1}{32}}\}$, where $F_\frac{1}{x}$ represent features extracted after $x \times$ downsampling. Then, we employ the Feature Pyramid Network (FPN)~\cite{lin2017feature} as the neck to aggregate these multi-scale features. The output feature map $F_{c_p}$ has a shape of $N_p \times C_p \times \frac{H_p}{8} \times \frac{W_p}{8}$, where $H_p$ and $W_p$ represent the input resolution of the image, and $C_p$ and $N_p$ indicate the number of channels and the number of surround images, respectively.

\subsubsection{Projection and Interpolation}
For image-related occupancy prediction, a key step is to transform image features from 2D image plane to 3D volume space. Most existing methods use monocular depth estimation~\cite{wang2023openoccupancy, huang2021bevdet, yu2023flashocc} or deformable attention~\cite{wei2023surroundocc, tian2024occ3d, wang2024panoocc, zhang2024occfusion}. However, monocular depth estimation is inherently an ill-posed problem~\cite{huang2023detecting, li2023bevstereo}, while deformable attention imposes a significant computational burden~\cite{harley2023simple}. Given these limitations, we use a simple yet effective projection and sample approach similar to Harley \textit{et al.}~\cite{harley2019learning}. Concretely, we first predefine a 3D voxel grid with a shape of $Z \times \frac{H}{8} \times \frac{W}{8}$, where $Z$ represents the number of voxels along the $z$-axis. The center point of each voxel is then projected onto the image feature plane, and only points that fall within both the image feature plane and the camera's field of view are retained. Next, the sub-pixel projection positions of the retained points are bilinearly interpolated to generate the image features corresponding to each voxel. For voxels located in the overlapping viewing regions of the surround cameras, we randomly select the image feature from one of the two corresponding cameras to obtain the final feature for each voxel. The output camera's voxel features can be denoted as $F_c \in \mathbb{R}^{C \times Z \times \frac{H}{8} \times \frac{W}{8}}$.

\subsubsection{BEV Encoder}
Given the fused feature $F_f$, we further refine $F_f$ by passing it through three blocks of ResNet18~\cite{he2016deep}, resulting in two feature maps, $F_{f0}$ and $F_{f2}$, extracted at two scales from the first and last blocks respectively. Then, similar to FPN~\cite{lin2017feature}, we apply bilinear upsampling to $F_{f2}$ and concatenate it with $F_{f0}$ along the feature dimension. Finally, we fuse the features of different scales using a convolution block. The output refined BEV features can be denoted as $F_r \in \mathbb{R}^{C_r \times \frac{H}{8} \times \frac{W}{8}}$.

\subsection{BVRE}
\label{sec:bvre}
As shown in Figure~\ref{fig:flops}, due to the sparsity of point clouds, the increase in computational cost resulting from expanding the processing range of the point cloud is very small compared to increasing the image resolution. Therefore, we extend the point cloud range to provide more 3D spatial context, compensating for the information loss caused by lower image resolution. However, arbitrarily setting the $XY$ range of the point cloud or voxel resolution may result in misalignment between the BEV features and the occupancy ground truth. Assuming that the voxel resolution of the occupancy grid is $res_o$, the expanded point cloud data in BEV covers a rectangular region with $X$ ranging from $-x$ to $x$ and $Y$ ranging from $-y$ to $y$, and has a voxel resolution of $res_p$. Thus, $x$ and $y$ must be integer multiples of $res_o$, and $res_o$ must also be an integer multiple of $res_p$. Otherwise, the spatial resolution represented by each feature vector along the $z$-axis in the BEV representation will not be equal to $res_o$. To avoid complex manual design, we adopt coordinate transformation and interpolation, as illustrated in the purple part of Figure~\ref{Fig:overview}, which can be formulated as follows:
\begin{equation}
    F_{occ} = {\rm GridSample}(F_{r}, {\rm Norm}(T_{o2l} \times P_o))
\end{equation}
where $P_o$ is a set of predefined points in the $XY$ plane of the occupancy annotation coordinate system, each point is located at the center of an occupancy grid in the $XY$ plane. $T_{o2l}$ is the transformation matrix from occupancy coordinates to LiDAR coordinates. The function of ${\rm Norm}$ is to scale the coordinate values to a range from -1 to 1.

\subsection{Detachable Auxiliary Detection Head}
\label{sec:det head}

To fully exploit the inherent geometry and structure information in point cloud features, we employ a detachable auxiliary detection head to further supervise the feature fusion process, thereby enhancing the information about object boundaries and scene structures in the fused features. Meanwhile, the auxiliary detection task is related to the occupancy prediction task, thus providing multiple regularization effects during optimization. For simplicity, we use the one-stage CenterHead introduced in CenterPoint~\cite{yin2021center} as the auxiliary detection head. Given the refined BEV feature $F_r$, we utilize two convolutional layers to generate a keypoint heatmap $HM = {p_{xy}}$ from the BEV perspective, and apply a Gaussian kernel~\cite{law2018cornernet, zhou2019objects} to map the center points of all ground truth 3D boxes onto a target heatmap $T$. The training objective is formulated as a focal loss~\cite{ross2017focal} based on Gaussian heatmaps:
\begin{equation}
    \mathcal{L}_{cls} = \frac{-1}{N} \sum_{ij}
    \begin{cases}
        (1 - p_{ij})^{\alpha} \log(p_{ij}) & \!\text{if}\ y_{ij} = 1 \\
        (1 - y_{ij})^{\beta} (p_{ij})^{\alpha} \log(1 - p_{ij}) & \text{otherwise}
    \end{cases}
\end{equation}
where $p_{ij}$ and $y_{ij}$ represent the predicted score and ground truth of the heatmap at location $(i, j)$, respectively. $N$ is the number of objects in a point cloud, and $\alpha$ and $\beta$ are hyper-parameters of the focal loss~\cite{ross2017focal}. Similarly, for the offset of the center point, as well as the height-above-ground, 3D dimensions and yaw of the 3D bounding box, we use separate convolutional layers to predict each of these parameters, and then apply the L1 loss for supervision:
\begin{equation}
    \mathcal{L}_{loc} = \frac{1}{N}\sum_{k=1}^{N} \left|\hat s_{k} - s_k\right|
\label{eq:size_loss}
\end{equation}
where $\hat s_{k}$ and $s_k$ represent the predicted value and the ground truth respectively. The above two tasks enable the network to perceive object boundaries more precisely, which in turn facilitates achieving more precise occupancy predictions. The total loss for the auxiliary detection head can be denoted as:
\begin{equation}
    \mathcal{L}_{det} = \mathcal{L}_{cls} + \lambda_l \mathcal{L}_{loc}
\end{equation}

\subsection{Overall Objective Function}
\label{sec:total loss}

Most existing methods rely on complex combinations of loss functions~\cite{wang2023openoccupancy, wang2024panoocc, li2023fb, yu2024panoptic, shi2024occupancy, wu2024deep, ming2024occfusion, zhang2024occfusion, pan2024co}, such as a combination of several or all of the focal loss~\cite{ross2017focal}, the scene-class affinity loss~\cite{cao2022monoscene}, the dice loss, the lovasz-softmax loss~\cite{berman2018lovasz}, the depth loss~\cite{li2023bevdepth}, etc., to achieve the expected performance. In contrast, our approach utilizes a detachable 3D object detection branch for auxiliary supervision and requires only a simple cross-entropy loss $\mathcal{L}_{ce}$ for occupancy prediction. The total loss for our framework can be defined as:
\begin{equation}
    \mathcal{L}_{total} = \mathcal{L}_{ce} + \lambda \mathcal{L}_{det}
\end{equation}

\section{Experiments}
\label{sec:experiments}

\definecolor{nothers}{RGB}{0, 0, 0}
\definecolor{nbarrier}{RGB}{255, 120, 50}
\definecolor{nbicycle}{RGB}{255, 192, 203}
\definecolor{nbus}{RGB}{255, 255, 0}
\definecolor{ncar}{RGB}{0, 150, 245}
\definecolor{nconstruct}{RGB}{0, 255, 255}
\definecolor{nmotor}{RGB}{200, 180, 0}
\definecolor{npedestrian}{RGB}{255, 0, 0}
\definecolor{ntraffic}{RGB}{255, 240, 150}
\definecolor{ntrailer}{RGB}{135, 60, 0}
\definecolor{ntruck}{RGB}{160, 32, 240}
\definecolor{ndriveable}{RGB}{255, 0, 255}
\definecolor{nother}{RGB}{139, 137, 137}
\definecolor{nsidewalk}{RGB}{75, 0, 75}
\definecolor{nterrain}{RGB}{150, 240, 80}
\definecolor{nmanmade}{RGB}{230, 230, 250}
\definecolor{nvegetation}{RGB}{0, 175, 0}

\begin{table*}[t!]
\caption{\textbf{3D occupancy prediction performance on Occ3D-nuScenes validation set.} The camera visible mask is used during the training phase. (xf) means use x frames for temporal fusion. * denotes the use of exponential moving average hook. C, L, and R represent camera, LiDAR, and radar, respectively. Image Size and 2D Backbone represent the input image resolution and the image feature extractor, respectively. FPS for SDGOCC and ALOcc are sourced directly from their original publications, evaluated on NVIDIA RTX 4090. All other FPS results are obtained through our own benchmarking on identical hardware (NVIDIA RTX 4090) using the PyTorch FP32 backend. The best results are shown in \textbf{bold}.}
\setlength{\tabcolsep}{0.0035\linewidth}
\centering
\resizebox{.98\linewidth}{!}{
\begin{tabular}{l | c | c | c c | c | c c c c c c c c c c c c c c c c c}
    \toprule
    Method
    & mIoU
    & Modality
    & Image Size & 2D Backbone
    & FPS
    & \rotatebox{90}{\textcolor{nothers}{$\blacksquare$} others}
    & \rotatebox{90}{\textcolor{nbarrier}{$\blacksquare$} barrier}
    & \rotatebox{90}{\textcolor{nbicycle}{$\blacksquare$} bicycle}
    & \rotatebox{90}{\textcolor{nbus}{$\blacksquare$} bus}
    & \rotatebox{90}{\textcolor{ncar}{$\blacksquare$} car}
    & \rotatebox{90}{\textcolor{nconstruct}{$\blacksquare$} const. veh.}
    & \rotatebox{90}{\textcolor{nmotor}{$\blacksquare$} motorcycle}
    & \rotatebox{90}{\textcolor{npedestrian}{$\blacksquare$} pedestrian}
    & \rotatebox{90}{\textcolor{ntraffic}{$\blacksquare$} traffic cone}
    & \rotatebox{90}{\textcolor{ntrailer}{$\blacksquare$} trailer}
    & \rotatebox{90}{\textcolor{ntruck}{$\blacksquare$} truck}
    & \rotatebox{90}{\textcolor{ndriveable}{$\blacksquare$} drive. suf.}
    & \rotatebox{90}{\textcolor{nother}{$\blacksquare$} other flat}
    & \rotatebox{90}{\textcolor{nsidewalk}{$\blacksquare$} sidewalk}
    & \rotatebox{90}{\textcolor{nterrain}{$\blacksquare$} terrain}
    & \rotatebox{90}{\textcolor{nmanmade}{$\blacksquare$} manmade}
    & \rotatebox{90}{\textcolor{nvegetation}{$\blacksquare$} vegetation}
    \\
    \midrule
    BEVDetOcc(2F)~\cite{huang2021bevdet} & 42.02 & C & 512 $\times$ 1408 & Swin-B & 1.2 & 12.2 & 49.6 & 25.1 & 52.0 & 54.5 & 27.9 & 28.0 & 28.9 & 27.2 & 36.4 & 42.2 & 82.3 & 43.3 & 54.6 & 57.9 & 48.6 & 43.6 \\
    PanoOcc(4F)~\cite{wang2024panoocc} & 42.13 & C & 864 $\times$ 1600 & R101-DCN & 3.0 & 11.7 & 50.5 & 29.6 & 49.4 & 55.5 & 23.3 & 33.3 & 30.6 & 31.0 & 34.4 & 42.6 & 83.3 & 44.2 & 54.4 & 56.0 & 45.9 & 40.4 \\
    FB-OCC(16F)~\cite{li2023fb} & 48.90 & C & 960 $\times$ 1760 & VoVNet-99 & - & 14.3 & 57.0 & 38.3 & 57.7 & 62.1 & 34.4 & 39.4 & 38.8 & 39.4 & 42.9 & 50.0 & 86.0 & 50.2 & 60.1 & 62.5 & 52.4 & 45.7 \\
    FlashOcc(2F)*~\cite{yu2023flashocc} & 43.52 & C & 512 $\times$ 1408 & Swin-B & 1.2 & 13.4 & 51.1 & 27.7 & 51.6 & 56.2 & 27.3 & 30.0 & 29.9 & 29.8 & 37.8 & 43.5 & 83.8 & 46.6 & 56.2 & 59.6 & 50.8 & 44.7 \\
    COTR(8F)~\cite{ma2024cotr} & 46.21 & C & 512 $\times$ 1408 & Swin-B & 0.3 & 14.9 & 53.3 & 35.2 & 50.8 & 57.3 & 35.4 & 34.1 & 33.5 & 37.1 & 39.0 & 45.0 & 84.5 & 48.7 & 57.6 & 61.1 & 51.6 & 46.7 \\
    STCOcc(16F)~\cite{liao2025stcocc} & 45.0 & C & 512 $\times$ 1408 & R50 & 1.4 & 15.2 & 52.3 & 32.2 & 50.5& 56.5 & 31.7 & 33.9 & 33.4 & 33.8 & 38.9 & 44.9 & 83.9 & 47.4 & 57.1 & 60.1 & 50.6 & 42.7 \\
    ALOcc(16F)~\cite{chen2024alocc} & 50.6 & C & 512 $\times$ 1408 & Swin-B & 1.5 & \textbf{17.0} & 58.3 & 39.7 & 56.6 & 63.2 & 33.2 & 41.3 & 40.3 & 40.8 & 43.8 & 51.0 & \textbf{87.0} & \textbf{52.7} & \textbf{62.0} & \textbf{65.2} & 57.7 & 50.9 \\
    RadOcc~\cite{zhang2024radocc} & 49.38 & C+L & 512 $\times$ 1408 & Swin-B & - & 10.9 & 58.2 & 25.0 & 57.9 & 62.9 & 34.0 & 33.5 & 50.1 & 32.1 & 48.9 & 52.1 & 82.9 & 42.7 & 55.3 & 58.3 & 68.6 & 66.0 \\
    OccFusion~\cite{ming2024occfusion} & 46.67 & C+L+R & 900 $\times$ 1600 & R101-DCN & 1.0 & 12.4 & 50.3 & 31.5 & 57.6 & 58.8 & 34.0 & 41.0 & 47.2 & 29.7 & 42.0 & 48.0 & 78.4 & 35.7 & 47.3 & 52.7 & 63.5 & 63.3 \\
    OccFusion~\cite{zhang2024occfusion} & 48.74 & C+L & 900 $\times$ 1600 & R101 & - & 12.4 & 51.8 & 33.0 & 54.6 & 57.7 & 34.0 & 43.0 & 48.4 & 35.5 & 41.2 & 48.6 & 83.0 & 44.7 & 57.1 & 60.0 & 62.5 & 61.3 \\
    SDGOCC~\cite{duan2025sdgocc} & 51.66 & C+L & - & R50 & 7.5 & 13.2 & 57.8 & 24.3 & 60.3 & 64.3 & 36.2 & 39.4 & 52.4 & 35.8 & \textbf{50.9} & 53.7 & 84.6 & 47.5 & 58.0 & 61.6 & \textbf{70.7} & 67.7 \\
    \midrule
    DAOcc (Ours) & \textbf{54.33} & C+L & \textbf{256} $\times$ \textbf{704} & \textbf{R50} & \textbf{7.8} & 13.0 & \textbf{60.7} & \textbf{39.8} & \textbf{64.0} & \textbf{66.5} & \textbf{36.3} & \textbf{49.0} & \textbf{60.1} & \textbf{44.3} & 50.7 & \textbf{55.9} & 82.9 & 44.6 & 56.8 & 60.6 & 70.1 & \textbf{68.3} \\
    \bottomrule
\end{tabular}}
\vspace{-8pt}
\label{tab:occ3d_nus_mask_w_fps}
\end{table*}

\begin{table}[t]
    \caption{Performance of DAOCC across different frameworks, precisions, and hardware, reported in mIoU and FPS. The reported mIoU is the result on Occ3D-nuScenes validation set with camera mask used for training.}
    \small
    \begin{center}
    \setlength{\tabcolsep}{0.015\linewidth}
    \scalebox{0.95}{
        \begin{tabular}{l | c c c | c c}
		    \toprule
            Method & Framework & Precision & Hardware & mIoU & FPS \\
            \midrule
            DAOcc & PyTorch & FP32 & RTX4090 & 54.33 & 7.8 \\
            DAOcc & TensorRT & FP16 & RTX4090 & 54.20 & 104.9 \\
            DAOcc & TensorRT & FP16+INT8 & Orin & 53.70 & 20.0 \\
            \bottomrule
        \end{tabular}
    }
    \end{center}
\label{tab:deploy}
\end{table}

\begin{table*}[t!]
\caption{\textbf{3D occupancy prediction performance on Occ3D-nuScenes validation set.} The camera visible mask is \textbf{not} used during the training phase. (xf) means use x frames for temporal fusion. C and L represent camera and LiDAR, respectively. The best results are shown in \textbf{bold}.}
\centering
\resizebox{.9\linewidth}{!}{
\begin{tabular}{l | c | c | c c | c c | c | c c c}
    \toprule
    \multirow{2}{*}{Method} 
    & \multirow{2}{*}{mIoU} 
    & \multirow{2}{*}{RayIoU} 
    & \multicolumn{2}{c |}{Modality} 
    & \multirow{2}{*}{Image Size} & \multirow{2}{*}{2D Backbone} 
    & \multirow{2}{*}{FPS}
    & \multirow{2}{*}{RayIoU\textsubscript{1m}} 
    & \multirow{2}{*}{RayIoU\textsubscript{2m}} 
    & \multirow{2}{*}{RayIoU\textsubscript{4m}} 
    \\
    & & & \textit{train} & \textit{val} & & & & & & \\
    \midrule
    BEVFormer(4F)~\cite{li2022bevformer} & 39.2 & 32.4 & - & C & 900 $\times$ 1600 & R101 & - & 26.1 & 32.9 & 38.0 \\
    RenderOcc~\cite{pan2023renderocc} & 24.4 & 19.5 & - & C & 512 $\times$ 1408 & Swin-B & - & 13.4 & 19.6 & 25.5 \\
    SimpleOcc~\cite{gan2024comprehensive} & 31.8 & 22.5 & - & C & 336 $\times$ 672 & R101 & - & 17.0 & 22.7 & 27.9 \\
    BEVDetOcc(8F)~\cite{huang2021bevdet} & 39.3 & 32.6 & C+L & C & 384 $\times$ 704 & R50 & - & 26.6 & 33.1 & 38.2 \\
    FB-Occ(16F)~\cite{li2023fb} & 39.1 & 33.5 & C+L & C & 256 $\times$ 704 & R50 & - & 26.7 & 34.1 & 39.7 \\
    SparseOcc(16F)~\cite{liu2023fully} & 30.9 & 36.1 & - & C & 256 $\times$ 704 & R50 & - & 30.2 & 36.8 & 41.2 \\
    Panoptic-FlashOcc~\cite{yu2024panoptic} & 29.4 & 35.2 & C+L & C & 256 $\times$ 704 & R50 & - & 29.4 & 36.0 & 40.1 \\
    Panoptic-FlashOcc(8F)~\cite{yu2024panoptic} & 31.6 & 38.5 & C+L & C & 256 $\times$ 704 & R50 & - & 32.8 & 39.3 & 43.4 \\
    STCOcc(16F)~\cite{liao2025stcocc} & - & 42.1 & C+L & C & 512 $\times$ 1408 & R50 & 1.4 & 36.9 & 42.8 & 46.7 \\
    ALOcc(16F)~\cite{chen2024alocc} & 38.0 & 43.7 & C+L & C & 256 $\times$ 704 & R50 & 6.0 & 37.8 & 44.7 & 48.8 \\
    \midrule
    DAOcc (Ours) & \textbf{48.3} & \textbf{48.4} & C+L & C+L & 256 $\times$ 704 & R50 & \textbf{7.8} & \textbf{44.5} & \textbf{48.9} & \textbf{51.9} \\
    \bottomrule
\end{tabular}}
\vspace{-8pt}
\label{tab:occ3d_nus_wo_mask}
\end{table*}



\begin{table}[t]
    \caption{\textbf{Performance comparisons between different BEV ranges and voxel sizes.} H-Range represents the range along the z-axis. In all experiments, H-Range is voxelized into 40 voxels.}
    \small
    \begin{center}
    \scalebox{0.85}{
    \setlength{\tabcolsep}{0.01\linewidth}
        \begin{tabular}{c | c | c | c | c}
		    \toprule
              & BEV View Range (m) & BEV Voxel Size (m) & H-Range (m) & mIoU \\
            \midrule
            (a) & $\left[-41.4, 41.4\right]$ & 0.075 $\times$ 0.075 & [-6, 4] & 50.57 \\
            (b) & $\left[-41.4, 41.4\right]$ & 0.075 $\times$ 0.075 & [-4, 2] & 50.27 \\
            (c) & $\left[-41.4, 41.4\right]$ & 0.075 $\times$ 0.075 & [-5, 3] & 50.76 \\
            (d) & $\left[-45.6, 45.6\right]$ & 0.075 $\times$ 0.075 & [-5, 3] & 50.98 \\
            (e) & $\left[-45.6, 45.6\right]$ & 0.1 $\times$ 0.1 & [-5, 3] & 49.75 \\
            (f) & $\left[-51.2, 51.2\right]$ & 0.1 $\times$ 0.1 & [-5, 3] & 49.98 \\
            (g) & $\left[-54.0, 54.0\right]$ & 0.075 $\times$ 0.075 & [-5, 3] & \textbf{51.26} \\
            \bottomrule
        \end{tabular}
    }
    \end{center}
\vspace{-8pt}
\label{tab:bvre}
\end{table}

\definecolor{nothers}{RGB}{0, 0, 0}
\definecolor{nbarrier}{RGB}{255, 120, 50}
\definecolor{nbicycle}{RGB}{255, 192, 203}
\definecolor{nbus}{RGB}{255, 255, 0}
\definecolor{ncar}{RGB}{0, 150, 245}
\definecolor{nconstruct}{RGB}{0, 255, 255}
\definecolor{nmotor}{RGB}{200, 180, 0}
\definecolor{npedestrian}{RGB}{255, 0, 0}
\definecolor{ntraffic}{RGB}{255, 240, 150}
\definecolor{ntrailer}{RGB}{135, 60, 0}
\definecolor{ntruck}{RGB}{160, 32, 240}
\definecolor{ndriveable}{RGB}{255, 0, 255}
\definecolor{nother}{RGB}{139, 137, 137}
\definecolor{nsidewalk}{RGB}{75, 0, 75}
\definecolor{nterrain}{RGB}{150, 240, 80}
\definecolor{nmanmade}{RGB}{213, 213, 213}
\definecolor{nvegetation}{RGB}{0, 175, 0}
\definecolor{diffdarkgreen}{rgb}{0.0, 0.5, 0.0}
\definecolor{diffdarkred}{rgb}{0.5, 0.0, 0.0}

\begin{table*}[t!]
\caption{\textbf{Ablation study on the Auxiliary Detection Head.} ADH denotes the Auxiliary Detection Head. The categories highlighted in light gray are those supervised during detection training, while the others remain unsupervised.}
\setlength{\tabcolsep}{0.0035\linewidth}
\centering
\resizebox{.98\linewidth}{!}{
\begin{tabular}{c | c | c | c c c c c c c c c c c c c c c c c}
    \toprule
    \makecell{camera \\ mask} & ADH & mIoU
    & \rotatebox{90}{\textcolor{nothers}{$\blacksquare$} others}
    & \cellcolor{gray!20}\rotatebox{90}{\textcolor{nbarrier}{$\blacksquare$} barrier}
    & \cellcolor{gray!20}\rotatebox{90}{\textcolor{nbicycle}{$\blacksquare$} bicycle}
    & \cellcolor{gray!20}\rotatebox{90}{\textcolor{nbus}{$\blacksquare$} bus}
    & \cellcolor{gray!20}\rotatebox{90}{\textcolor{ncar}{$\blacksquare$} car}
    & \cellcolor{gray!20}\rotatebox{90}{\textcolor{nconstruct}{$\blacksquare$} const. veh.}
    & \cellcolor{gray!20}\rotatebox{90}{\textcolor{nmotor}{$\blacksquare$} motorcycle}
    & \cellcolor{gray!20}\rotatebox{90}{\textcolor{npedestrian}{$\blacksquare$} pedestrian}
    & \cellcolor{gray!20}\rotatebox{90}{\textcolor{ntraffic}{$\blacksquare$} traffic cone}
    & \cellcolor{gray!20}\rotatebox{90}{\textcolor{ntrailer}{$\blacksquare$} trailer}
    & \cellcolor{gray!20}\rotatebox{90}{\textcolor{ntruck}{$\blacksquare$} truck}
    & \rotatebox{90}{\textcolor{ndriveable}{$\blacksquare$} drive. suf.}
    & \rotatebox{90}{\textcolor{nother}{$\blacksquare$} other flat}
    & \rotatebox{90}{\textcolor{nsidewalk}{$\blacksquare$} sidewalk}
    & \rotatebox{90}{\textcolor{nterrain}{$\blacksquare$} terrain}
    & \rotatebox{90}{\textcolor{nmanmade}{$\blacksquare$} manmade}
    & \rotatebox{90}{\textcolor{nvegetation}{$\blacksquare$} vegetation} \\
    \midrule
    \checkmark & & 51.26 & \textbf{13.01} & 56.94 & 20.6 & 61.49 & 65.51 & 35.08 & 38.89 & 55.44 & 36.65 & 48.46 & 53.82 & \textbf{83.22} & \textbf{45.08} & \textbf{57.35} & \textbf{60.63} & \textbf{70.83} & \textbf{68.47} \\
    \checkmark & \checkmark & \makecell{\textbf{52.82} \\ (\textcolor{diffdarkgreen}{+1.56})} & \makecell{10.93 \\ (\textcolor{diffdarkred}{-2.08})} & \makecell{\textbf{58.25} \\ (\textcolor{diffdarkgreen}{+1.31})} & \makecell{\textbf{36.35} \\ (\textcolor{diffdarkgreen}{+15.75})} & \makecell{\textbf{62.49} \\ (\textcolor{diffdarkgreen}{+1})} & \makecell{\textbf{65.91} \\ (\textcolor{diffdarkgreen}{+0.4})} & \makecell{\textbf{35.45} \\ (\textcolor{diffdarkgreen}{+0.37})} & \makecell{\textbf{45.48} \\ (\textcolor{diffdarkgreen}{+6.59})} & \makecell{\textbf{58.27} \\ (\textcolor{diffdarkgreen}{+2.83})} & \makecell{\textbf{41.98} \\ (\textcolor{diffdarkgreen}{+5.33})} & \makecell{\textbf{48.67} \\ (\textcolor{diffdarkgreen}{+0.21})} & \makecell{\textbf{55.04} \\ (\textcolor{diffdarkgreen}{+1.22})} & \makecell{82.49 \\ (\textcolor{diffdarkred}{-0.73})} & \makecell{42.18 \\ (\textcolor{diffdarkred}{-2.9})} & \makecell{56.16 \\ (\textcolor{diffdarkred}{-1.19})} & \makecell{59.88 \\ (\textcolor{diffdarkred}{-0.75})} & \makecell{70.41 \\ (\textcolor{diffdarkred}{-0.42})} & \makecell{67.94 \\ (\textcolor{diffdarkred}{-0.53})} \\
    \bottomrule
\end{tabular}}
\vspace{-8pt}
\label{tab:ablation_det_detail}
\end{table*}

\begin{figure}[t]
    \centering
    \includegraphics[width=\columnwidth]{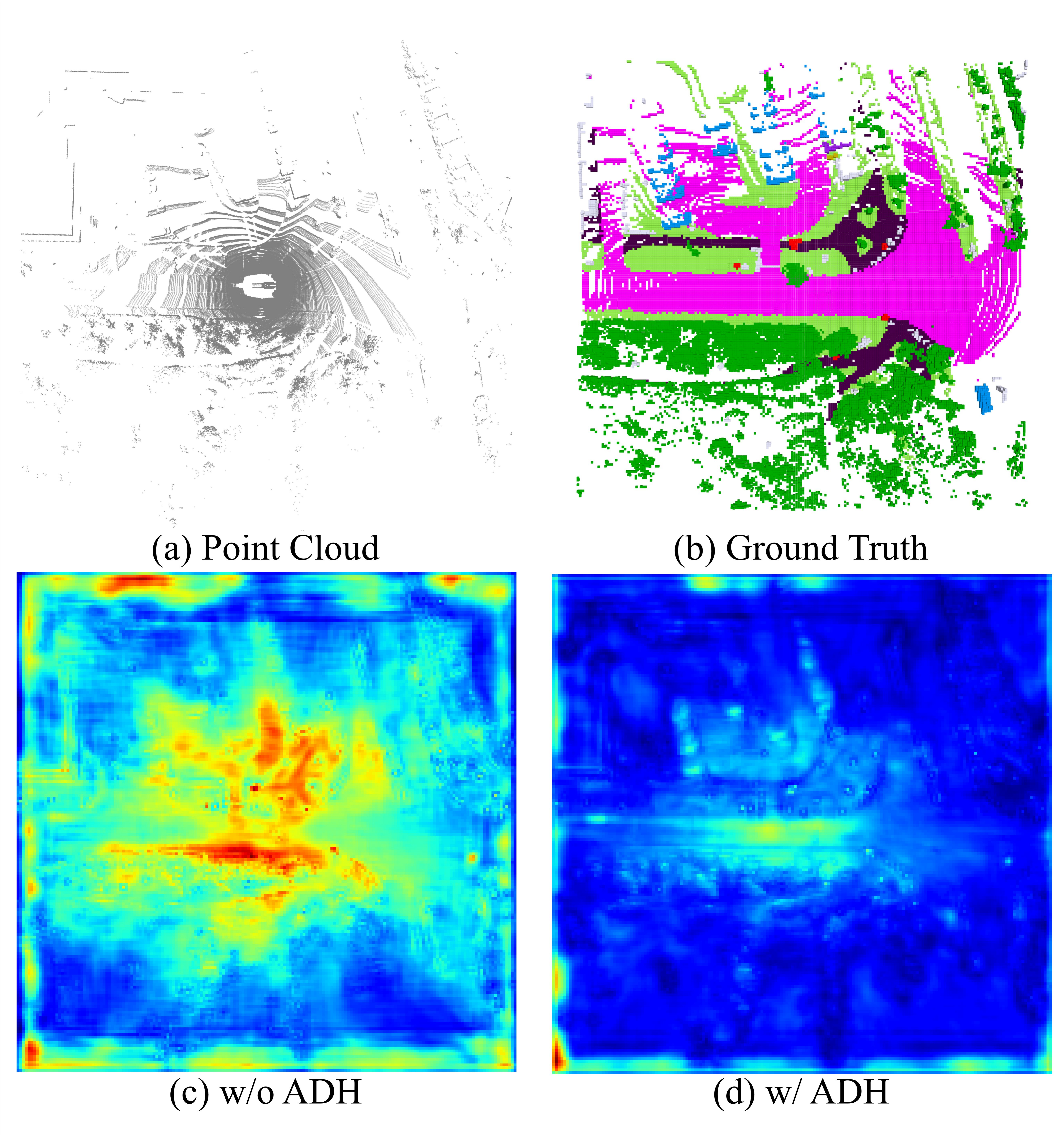}
    \caption{\textbf{Visualization of BEV feature map distributions under different supervision configurations.} ADH denotes the Auxiliary Detection Head. Subfigures (a) and (b) show the raw input point cloud data and the corresponding 3D occupancy ground-truth labels. Subfigures (c) and (d) compare the BEV features without and with ADH supervision. The incorporation of ADH supervision enhances the BEV feature representations in two aspects: (1) scene boundaries become more clearly delineated; and (2) foreground objects (e.g., vehicles in the upper parking lot) appear more distinguishable.}
    \vspace{-8pt}
    \label{fig:heatmap}
\end{figure}

\begin{figure}[t]
    \centering
    \includegraphics[width=0.94\columnwidth]{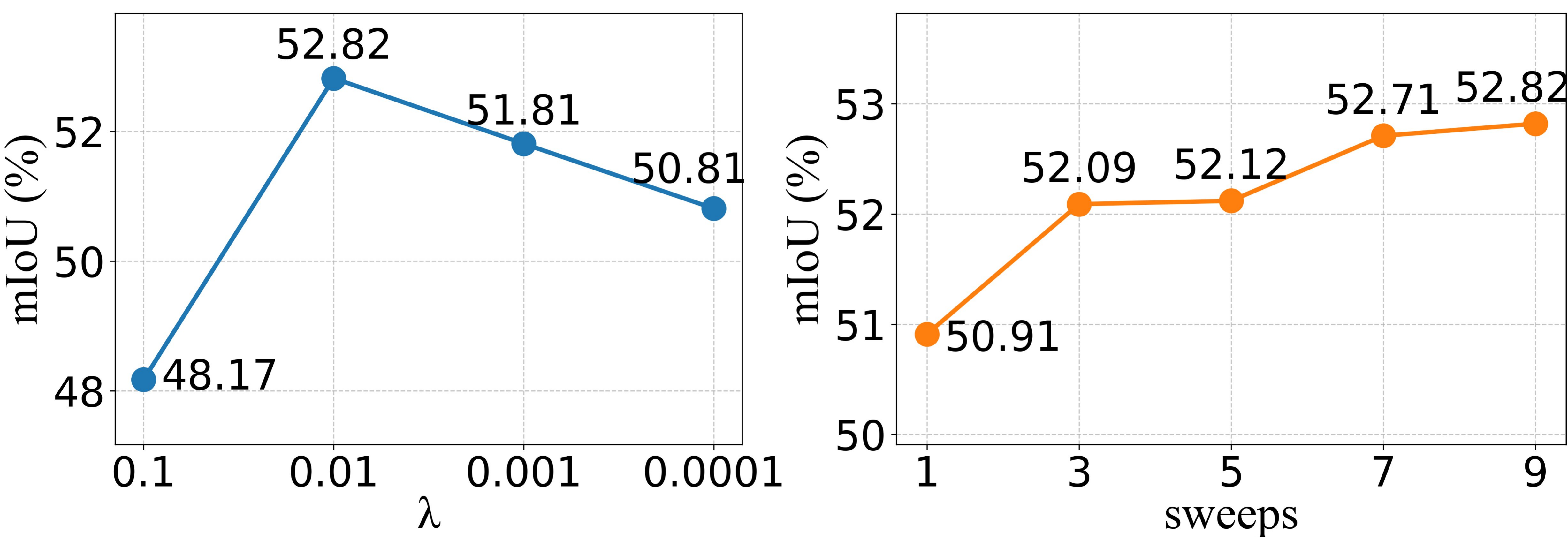}
    \caption{\textbf{Effect of loss weight $\lambda$ and LiDAR sweeps.}}
    \vspace{-8pt}
    \label{fig:lambda}
\end{figure}

\begin{table}[t]
    \caption{\textbf{Ablation for achieving final performance.} EP15, EP24, and EP6 represent training for 15 epochs, 24 epochs, and 6 epochs, respectively.}
    \small
    \begin{center}
    \scalebox{0.95}{
        \begin{tabular}{c | c}
		    \toprule
            Scheduler & mIoU \\
            \midrule
            EP15 & 52.82 \\
            EP24 & 53.39 \\
            CBGS(EP6) & 53.82 \\
            CBGS(EP6)+LR$\times$2 & \textbf{54.33} \\
            \bottomrule
        \end{tabular}
    }
    \end{center}
\vspace{-8pt}
\label{tab:performance}
\end{table}

\begin{table*}[t]
\caption{\textbf{3D occupancy prediction performance on Occ3D-Waymo validation set.} * denotes the use of exponential moving average hook. C and L represent camera and LiDAR, respectively. GO stands for general object and Cons. Cone represents the construction cone. The best results are shown in \textbf{bold}.}
\setlength{\tabcolsep}{0.005\linewidth}
\centering
\resizebox{.98\linewidth}{!}{
    \begin{tabular}{l| c | c | c c | c c c c c c c c c c c c c c c}
        \toprule
        Method
        & mIoU
        & Modality
        & Image Size & 2D Backbone
        & \rotatebox{90}{GO}
        & \rotatebox{90}{vehicle}
        & \rotatebox{90}{bicyclist}
        & \rotatebox{90}{pedestrian}
        & \rotatebox{90}{sign}
        & \rotatebox{90}{traffic light}
        & \rotatebox{90}{pole}
        & \rotatebox{90}{Cons. Cone}
        & \rotatebox{90}{bicycle}
        & \rotatebox{90}{motorcycle}
        & \rotatebox{90}{building}
        & \rotatebox{90}{vegetation}
        & \rotatebox{90}{tree trunk}
        & \rotatebox{90}{road}
        & \rotatebox{90}{sidewalk}
        \\
        \midrule
        BEVDet~\cite{huang2021bevdet} & 9.88 & C & - & - & 0.13 & 13.06 & 2.17 & 10.15 & 7.80 & 5.85 & 4.62 & 0.94 & 1.49 & 0.0 & 7.27 & 10.06 & 2.35 & 48.15 & 34.12 \\
        TPVFormer~\cite{huang2023tri} & 16.76 & C & - & - & 3.89 & 17.86 & 12.03 & 5.67 & 13.64 & 8.49 & 8.90 & 9.95 & 14.79 & 0.32 & 13.82 & 11.44 & 5.8 & 73.3 & 51.49 \\
        BEVFormer~\cite{li2022bevformer} & 16.76 & C & - & - & 3.48 & 17.18 & 13.87 & 5.9 & 13.84 & 2.7 & 9.82 & 12.2 & 13.99 & 0.0 & 13.38 & 11.66 & 6.73 & 74.97 & 51.61 \\
        CTF-Occ~\cite{tian2024occ3d} & 18.73  & C & 640 $\times$ 960 & R101-DCN & 6.26 & 28.09 & 14.66 & 8.22 & 15.44 & 10.53 & 11.78 & 13.62 & 16.45 & 0.65 & 18.63 & 17.3 & 8.29 & 67.99 & 42.98 \\
        OPUS~\cite{wang2024opus} & 19.00 & C & - & - & 4.66 & 27.07 & 19.39 & 6.53 & 18.66 & 6.41 & 11.44 & 10.40 & 12.90 & 0.00 & 18.73 & 18.11 & 7.46 & 72.86 & 50.31 \\
        ODG~\cite{shi2025odg} & 21.35 & C & 640 $\times$ 960 & R50 & 5.09 & 31.34 & 22.4 & 19.06 & 15.24 & 6.09 & 12.51 & 12.77 & 13.59 & 0.00 & 21.49 & 17.89 & 8.37 & 78.19 & 56.28 \\
        LiDAR-Only~\cite{tian2024occ3d} & 29.74 & L & - & - & 1.01 & 57.41 & 35.31 & 20.33 & 11.7 & 13.01 & 36.21 & 7.81 & 0.13 & 0.0 & 57.83 & 54.71 & 27.07 & 69.15 & 54.47 \\
        BEVFormer-Fusion~\cite{tian2024occ3d} & 39.05 & C+L & - & - & 5.11 & 64.61 & 52.35 & 21.52 & 32.74 & 17.1 & 42.62 & 27.75 & 13.36 & 0.05 & \textbf{63.65} & 60.51 & 35.64 & \textbf{81.89} & \textbf{66.84} \\ 
        \midrule
        DAOcc (Ours) & 44.69 & C+L & 256 $\times$ 704 & R50 & 7.64 & 65.60 & 54.05 & \textbf{43.71} & 52.78 & 32.05 & 46.96 & 32.68 & 31.68 & \textbf{0.68} & 55.94 & 60.19 & 42.59 & 79.62 & 64.09 \\
        DAOcc* (Ours) & \textbf{45.13} & C+L & 256 $\times$ 704 & R50 & \textbf{8.15} & \textbf{65.90} & \textbf{54.40} & 43.36 & \textbf{52.86} & \textbf{32.93} & \textbf{47.54} & \textbf{33.54} & \textbf{32.29} & 0.67 & 57.49 & \textbf{60.56} & \textbf{42.86} & 79.78 & 64.69 \\
        \bottomrule
    \end{tabular}}
\vspace{-8pt}
\label{tab:occ3d-waymo}
\end{table*}

\begin{figure}[t]
    \centering
    \includegraphics[width=\columnwidth]{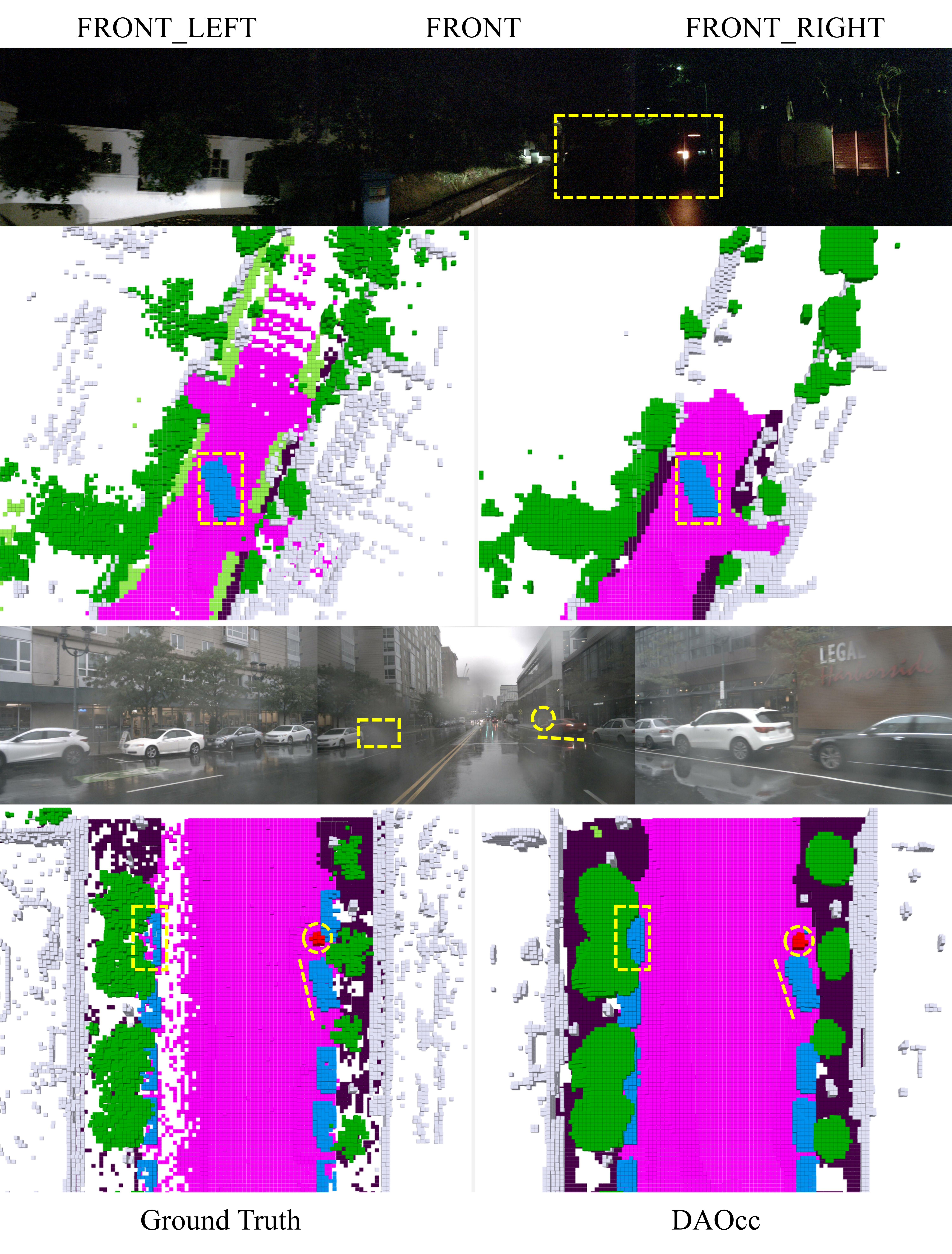}
    \caption{\textbf{Qualitative results of low-illumination and occluded scenarios on the Occ3D-nuScenes validation set.} The first two rows show nighttime scenes, and the last two rows depict rainy scenes. In each scene, yellow dotted markers indicate target objects that are difficult to identify. In rainy scenes, yellow dotted lines are used to align the position of the same object between the image view and its corresponding 3D occupancy in the BEV perspective.}
    \vspace{-8pt}
    \label{fig:robust}
\end{figure}

\begin{figure*}[t]
    \centering
    \includegraphics[width=0.88\textwidth]{fig/vis_4x3_compressed.pdf}
    \caption{\textbf{Qualitative visualizations on Occ3D-nuScenes validation set.} The camera visible mask is \textbf{NOT} used during model training and visualization.} 
    \label{Fig:occ3d_vis}
    \vspace{-8pt}
\end{figure*}

\subsection{Datasets and Evaluation Metrics}

\subsubsection{Occ3D-nuScenes}
Occ3D-nuScenes~\cite{tian2024occ3d} is a large-scale benchmark dataset for 3D occupancy prediction, which contains 700 scenes for training and 150 scenes for validation. Each scene lasts for 20 seconds, with annotations provided at a frequency of 2 Hz. The dataset covers a perception range of [-40m, 40m] in the $X$ and $Y$ directions, and [-1m, 5.4m] along the $z$-axis, all within the ego coordinate system. This range is discretized into voxels, each with a size of [0.4m, 0.4m, 0.4m]. Each occupied voxel is assigned one of 17 semantic classes, including 16 common classes and a general object class labeled as `others'. The dataset includes RGB images captured by six surrounding cameras and LiDAR point clouds, enabling dense voxel-wise annotations for 3D scene understanding. Additionally, the dataset provides visibility masks for both LiDAR and camera modes, which can be utilized for training purposes.

\subsubsection{Occ3D-Waymo}
Occ3D-Waymo~\cite{tian2024occ3d} is a large-scale benchmark dataset for 3D occupancy prediction, containing 798 training scenes and 202 validation scenes. It shares the same perception range and voxel size as Occ3D-nuScenes. Each occupied voxel is annotated with one of the 15 semantic classes, including a \textit{General Object (GO)} class. The dataset provides RGB images from five cameras (with raw resolutions of $1280 \times 1920$ for front, front-left, and front-right cameras, and $886 \times 1920$ for side-left and side-right cameras) and LiDAR point clouds, along with visibility masks for LiDAR and camera modalities to facilitate training.



\subsubsection{Evaluation metrics}
Following previous methods, we use the mean Intersection over Union (mIoU) as our evaluation metric when training models with the camera visible mask provided by Occ3D-nuScenes and Occ3D-Waymo, ensuring a fair comparison with existing approaches. When the camera visible mask is excluded from the training phase, we report an additional semantic segmentation metric, RayIoU~\cite{liu2023fully}, 
which is a recently introduced metric. 
RayIoU evaluates occupancy along query rays at three distance thresholds (1m, 2m, 4m), with the mean of these results serving as the final score.


\subsection{Implementation Details}

We use ResNet-50~\cite{he2016deep}, pre-trained on nuImages~\cite{caesar2020nuscenes} by MMDetection~\cite{chen2019mmdetection}, as our image backbone and adjust the input image resolution to 256×704. In the LiDAR branch, we aggregate LiDAR sweeps from 10 adjacent timestamps as input and initiate training from scratch. Data augmentation techniques, such as scaling, rotation, and flipping, are applied to the input images, and we also perform random flipping in the BEV space. The loss weights are configured as $\lambda_l=0.25$ and $\lambda=0.01$. We adopt the AdamW optimizer~\cite{loshchilov2017decoupled} with a cosine annealing learning rate scheduler, incorporating a warmup phase, and set the initial learning rate to 2e-4. Unless conducting ablation experiments, we train our DAOcc using CBGS~\cite{zhu2019class} for 6 epochs, the input point cloud range is set to [-54.0, -54.0, -5.0, 54.0, 54.0, 3.0] and the voxel size is set to [0.075, 0.075, 0.2]. All models are trained on 4 NVIDIA RTX4090 GPUs with a batch size of 4. Additionally, if the camera visible mask is not used during training, we randomly discard 80\% of empty voxels, consistent with Zhao \textit{et al.}~\cite{zhao3d}. Temporal information fusion and test-time augmentation are \textbf{not} used in any experiment. Unless otherwise stated, exponential moving average (EMA) for weight parameter is not used.


\subsection{Comparison with State-of-the-Art Methods}


\subsubsection{Comparison on Occ3D-nuScenes with camera mask}
As presented in Table~\ref{tab:occ3d_nus_mask_w_fps}, we compare our proposed DAOcc with both image-based and multi-modal methods on the Occ3D-nuScenes dataset. All results are either provided directly by their authors or obtained from their official code implementations. FPS for SDGOCC and ALOcc are sourced directly from their original publications, evaluated on NVIDIA RTX 4090. All other FPS results are obtained through our own benchmarking on identical hardware (NVIDIA RTX 4090) using the PyTorch FP32 backend. 
Our proposed DAOcc establishes a new state-of-the-art performance, achieving an impressive 54.33 mIoU with only a 2D ResNet50~\cite{he2016deep} backbone and a low input image resolution of 256×704, while maintaining a competitive inference speed of 7.8 FPS.
Compared with the latest state-of-the-art multi-modal method SDGOCC, DAOcc achieves an improvement of 2.67 mIoU and maintains a higher FPS (7.8 vs 7.5).
Furthermore, DAOcc performs strongly across nearly all foreground semantic categories, which are critical for ensuring autonomous driving safety.

As shown in Table~\ref{tab:deploy}, we further evaluate the performance and inference speed of DAOcc under different frameworks and hardware configurations. Unlike most LSS-based methods that rely on custom CUDA operators (e.g., BEV Pool~\cite{liu2023bevfusion}), the image branch of DAOcc does not require any custom operators when deployed with TensorRT and can be directly exported to the ONNX format, greatly simplifying the deployment process.
With TensorRT optimization, DAOcc achieves 104.9 FPS on an RTX 4090 while maintaining 54.2 mIoU, and 20.0 FPS on a Jetson AGX Orin (64GB) while maintaining 53.7 mIoU. Although quantization introduces a slight drop in mIoU performance, the resulting 53.7 mIoU still significantly surpasses all other methods reported in Table~\ref{tab:occ3d_nus_mask_w_fps}.





\subsubsection{Comparison on Occ3D-nuScenes without camera mask}
To the best of our knowledge, no existing multi-modal occupancy prediction methods have reported their performance using the RayIoU metric, so we mainly compare with image-based methods. It is worth noting that some image-based methods~\cite{huang2021bevdet,li2023fb,yu2024panoptic,liao2025stcocc,chen2024alocc}, such as the current state-of-the-art STCOcc and ALOcc, are based on Explicit Depth Supervision~\cite{li2023bevdepth} and leverage depth information derived from point clouds to supervise monocular depth estimation during training. As such, these methods can also be considered multi-modal in the training phase. 

As shown in Table~\ref{tab:occ3d_nus_wo_mask}, benefiting from the effective use of multi-modal inputs, especially point cloud data, our DAOcc achieves 48.3 mIoU and 48.4 RayIoU, significantly outperforming image-based methods by a large margin (surpassing ALOcc by 10.3 mIoU and 4.7 RayIoU), despite many of these methods employing multi-frame temporal fusion, which is an effective technique to improve RayIoU scores (e.g., +3.3 RayIoU for Panoptic-FlashOcc with 8-frame fusion vs. single-frame). Furthermore, when using the same 2D backbone (ResNet50) and image input resolution (256×704), although DAOcc additionally inputs point cloud data, its inference speed is still faster than the image-based method ALOcc (7.8 vs. 6.0).

\subsection{Ablation Studies}
\label{sec:ablation_studies}

Ablation studies are conducted on the validation set of Occ3D-nuScenes and Occ3D-Waymo, with the camera visible mask being used during training. Unless otherwise specified, all experiments presented in this section are trained for 15 epochs without employing CBGS~\cite{zhu2019class}.



\subsubsection{Effectiveness of the BVRE}
We propose the BVRE strategy to enrich spatial contextual information by expanding the perceptual range from a BEV (Bird’s Eye View) perspective. As shown in Table~\ref{tab:bvre}, a comparison of (c) and (d) indicates that, with the same voxel size of $0.075m \times 0.075m$ and the H-Range of [-5m, 3m], expanding the BEV range from [-41.4m, 41.4m] to [-45.6m, 45.6m] yields a performance improvement of 0.22 mIoU (from 50.76 to 50.98). Furthermore, the comparison of (c) and (g) shows that further expanding the BEV range to [-54.0m, 54.0m] with the same voxel size and H-Range increases the performance gain to 0.5 mIoU (from 50.76 to 51.26). This observation validates the value of spatial context information in a larger BEV range.

In addition, by comparing (d) and (e), the voxel size has a great impact on the model's performance. Specifically, increasing the voxel size from $0.075m \times 0.075m$ to $0.1m \times 0.1m$ while keeping the BEV range and H-Range the same leads to a decrease in mIoU from 50.98 to 49.75. Moreover, by analyzing the data in (a), (b), and (c), the H-Range of [-5m, 3m] is a relatively optimal setting, as it consistently achieves higher mIoU values compared to other H-Ranges. To achieve better performance, we adopt the setting (g) in DAOcc, which combines the largest BEV range of [-54.0m, 54.0m], the voxel size of $0.075 \times 0.075$ m, and the H-Range of [-5m, 3m].

\subsubsection{Ablation on the Auxiliary Detection Head}
To fully leverage the inherent geometry and structure information in point cloud features, we incorporate 3D object detection as auxiliary supervision. As shown in Table~\ref{tab:ablation_det_detail}, adding auxiliary 3D detection supervision leads to a 1.56 mIoU improvement when training with camera visible masks. Notably, this improvement originates entirely from foreground categories, which are highlighted in light gray in Table~\ref{tab:ablation_det_detail}. These categories correspond to the ones supervised by the auxiliary detection branch, demonstrating the effectiveness of our proposed supervision strategy.

Meanwhile, we observe that categories without detection supervision all exhibit varying degrees of performance degradation. We infer that this is because detection supervision emphasizes the localization of foreground objects, which strengthens the representation of foreground objects and thus weakens that of background categories. To mitigate this issue, we believe that additional BEV segmentation supervision could be introduced to balance the representation of foreground and background categories.


To further investigate how auxiliary 3D detection supervision affects the BEV feature distribution, we visualize intermediate features in Figure~\ref{fig:heatmap}. Subfigures (a) and (b) show the input point cloud and the corresponding ground-truth occupancy map, while (c) and (d) present the BEV features without and with auxiliary 3D detection supervision, respectively. The comparison reveals that incorporating auxiliary 3D detection enhances the preservation of structural scene details and significantly improves the distinguishability of foreground objects.





\subsubsection{Ablation on Loss Weight and LiDAR Sweeps}
Figure~\ref{fig:lambda} illustrates the impacts of two factors: the loss weight $\lambda$ and LiDAR sweeps on the mIoU.

To analyze the effect of loss weight $\lambda$, which balances the loss contributions between 3D occupancy prediction and 3D object detection, we vary its value across 0.1, 0.01, 0.001, and 0.0001. As shown in the left plot, mIoU rises sharply from 48.17 to 52.82 as $\lambda$ decreases from 0.1 to 0.01, then gradually declines to 50.81 as $\lambda$ is further reduced to 0.0001. This suggests that there exists an optimal $\lambda$ (around 0.01) that best balances the two loss terms for optimal mIoU.

The right plot explores the effect of LiDAR sweeps, referring to the number of point cloud frames aggregated over time. With sweeps increasing from 1 to 9, mIoU climbs steadily: starting at 50.91 (1 sweep), reaching 52.09 (3 sweeps), maintaining stability up to 5 sweeps, and then rising to 52.82 at 9 sweeps. Incorporating more sweeps generally enriches features from multi-frame data and improves mIoU performance, although the marginal gains diminish beyond a certain point.

\subsubsection{Ablation for Achieving Final Performance}
Table~\ref{tab:performance} presents the ablation experiments that contributed to the final performance of DAOcc. First, extending the training time from 15 epochs to 24 epochs yields an improvement of 0.57 mIoU. Next, using CBGS~\cite{zhu2019class} yields an additional 0.43 mIoU improvement. Since CBGS employs resampling, which increases the number of samples per epoch, we limit the training to 6 epochs to maintain a total training time comparable to that of the 24 epochs without CBGS for a fair comparison. Finally, increasing the learning rate can further improve performance.

\subsubsection{Comparison on Occ3D-Waymo with camera mask}
We further evaluate the robustness and generalization capability of DAOcc on the Occ3D-Waymo dataset, which remains a relatively underexplored benchmark for occupancy prediction. Following prior works~\cite{tian2024occ3d,wang2024opus,shi2025odg}, we train DAOcc using only 20\% of the training data and discard LiDAR sweeps by utilizing a single LiDAR frame for each input, ensuring a fair comparison.
As presented in Table~\ref{tab:occ3d-waymo}, DAOcc significantly outperforms all existing methods, surpassing the multi-modal baseline BEVFormer-Fusion by 5.64 mIoU. Compared to the nuScenes dataset, which features a single top-mounted LiDAR, the Waymo dataset provides denser point clouds captured from five LiDARs. This richer sensory input further amplifies the performance gap between multi-modal and image-only methods. For instance, the recently proposed image-based method ODG~\cite{shi2025odg} achieves an mIoU that is only 47.8\% of that achieved by DAOcc (21.35 vs. 44.69).
These results highlight the critical importance of high-quality point cloud data in improving 3D occupancy prediction, particularly in complex environments, while also validating the strong generalization ability of DAOcc across different datasets and sensing setups.


\subsubsection{Robustness}
We aim to apply state-of-the-art 3D occupancy prediction models to real-world scenarios by leveraging sparse LiDAR data more effectively within a multi-modal fusion framework. Although robust alignment and fusion of sensor modalities are not the primary focus of this work, we nonetheless evaluate DAOcc's robustness under challenging scenarios, as shown in Figure~\ref{fig:robust}.

In the nighttime scenario, under low illumination, the vehicle ahead (highlighted by the yellow dashed box) is barely observable in the camera image except for its lights. Even so, DAOcc successfully recognizes this vehicle and reconstructs a voxelized occupancy shape that closely matches the ground truth, demonstrating strong robustness in low-light environments.

In the rainy scenario, raindrops obscure the camera view. The vehicle parked on the left (yellow dashed box) is nearly invisible, yet DAOcc still recognizes it accurately. A pedestrian partially occluded by another vehicle on the right (yellow dashed circle) is also correctly reconstructed in voxel space. Although DAOcc misses the vehicle immediately behind this pedestrian, this target is almost indiscernible even by manual inspection, and the corresponding LiDAR data also does not show vehicle traces at this position. We thus attribute this miss to the intrinsic difficulty of the sample rather than a failure of our method.

The above results indicate that although DAOcc was not specifically designed for robustness, it already exhibits inherent robustness against interference due to the intrinsic generalization of data-driven learning paradigms. Based on this observation, we believe that increasing training data and designing specific data augmentation strategies for simulating sensor failures during training would further enhance DAOcc's robustness.




\subsubsection{Detection Performance}
Since our proposed DAOcc incorporates an auxiliary 3D detection branch into the occupancy prediction framework, we also evaluate the performance of this auxiliary branch. The auxiliary detection branch identifies 10 object categories, including cars, trucks, buses, trailers, construction vehicles, pedestrians, motorcycles, bicycles, traffic cones, and barriers. We train the auxiliary 3D detection branch on the nuScenes dataset~\cite{caesar2020nuscenes}, and the results are evaluated using the mean average precision (mAP) and nuScenes detection score (NDS) metrics provided by the nuScenes benchmark. DAOcc achieves 59.4 mAP and 64.3 NDS on the nuScenes validation set, and the APs for the 10 object categories (in the order listed above) are 85.8, 55.5, 66.1, 30.8, 22.0, 75.8, 67.4, 52.2, 69.6, 69.2.
These results show that DAOcc has the potential to be extended into a multi-task 3D perception framework.

\subsection{Visualization}
Figure~\ref{Fig:occ3d_vis} presents the visualizations of DAOcc on Occ3D-nuScenes validation set. The camera visible mask is not used during model training and visualization. The results demonstrate that DAOcc is capable of making a relatively complete prediction of the scene, accurately reproducing it in fine detail.

\section{Conclusion}

In this paper, we propose a novel multi-modal occupancy prediction framework, DAOcc, which aims to achieve superior performance while maintaining a deployment-friendly architecture with practical image backbones and input image resolutions. DAOcc achieves superior performance by fully leveraging the inherent geometry advantages of LiDAR data within a multi-modal framework, advantages largely overlooked by previous work.
Extensive experiments on the Occ3D-nuScenes and Occ3D-Waymo datasets demonstrate that DAOcc surpasses existing methods in both accuracy and inference speed.
Furthermore, due to its deployment-friendly design, DAOcc can be efficiently deployed and evaluated on edge devices. Notably, DAOcc runs in real-time on an AGX Orin device while maintaining state-of-the-art mIoU performance.


\setcounter{page}{1}
{
    \bibliographystyle{IEEEtran}
    \bibliography{main}
}



\end{document}